\documentclass[conference]{IEEEtran}
\IEEEoverridecommandlockouts

\usepackage{cite}
\usepackage{amsmath,amssymb,amsfonts}
\usepackage{algorithmic}
\usepackage{graphicx}
\usepackage{textcomp}
\usepackage{xcolor}
\usepackage{threeparttable}
\usepackage{booktabs}
\usepackage{multirow}
\usepackage{makecell}
\usepackage[table]{xcolor}
\usepackage{booktabs}
\usepackage{threeparttable}
\usepackage[table]{xcolor}
\usepackage{siunitx}
\usepackage{makecell}
\usepackage{array}
\usepackage{colortbl} 
\usepackage{siunitx}     
\usepackage[table]{xcolor} 
\usepackage{graphicx}
\usepackage{subcaption}
\usepackage{caption}

\usepackage[most]{tcolorbox}
\usepackage{amsmath}
\tcbuselibrary{breakable}
\usepackage{enumitem}

\def\BibTeX{{\rm B\kern-.05em{\sc i\kern-.025em b}\kern-.08em
    T\kern-.1667em\lower.7ex\hbox{E}\kern-.125emX}}

\newcommand{\equalcontrib}{\textsuperscript{$^{*}$}}
\newcommand{\projlead}{\textsuperscript{\ddag}}
\newcommand{\corrauthor}{\textsuperscript{\dag}}

\def\BibTeX{{\rm B\kern-.05em{\sc i\kern-.025em b}\kern-.08em
    T\kern-.1667em\lower.7ex\hbox{E}\kern-.125emX}}
    
\begin{document}

\title{ArchMap: Arch-Flattening and Knowledge-Guided Vision Language Model for Tooth Counting and Structured Dental Understanding%
\thanks{\noindent\parbox{\linewidth}{%
\equalcontrib~Co-first authors.\\
\projlead~Project Leader.\\
\corrauthor~Corresponding authors.}}%
}

\author{%
  \IEEEauthorblockN{%
    Bohan Zhang\textsuperscript{1}\equalcontrib,
    Yiyi Miao\textsuperscript{1,2}\equalcontrib\projlead,
    Taoyu Wu\textsuperscript{3,4}\equalcontrib,
    Tong Chen\textsuperscript{1,2},
    Ji Jiang\textsuperscript{5}, \\
    Zhuoxiao Li\textsuperscript{7},
    Zhe Tang\textsuperscript{6},
    Limin Yu\textsuperscript{3}\corrauthor,
    Jionglong Su\textsuperscript{1}\corrauthor}
  \IEEEauthorblockA{%
    \textsuperscript{1}School of AI and Advanced Computing, Xi'an Jiaotong-Liverpool University, China\\
    \textsuperscript{2}School of Electrical Engineering, Electronics and Computer Science, University of Liverpool, United Kingdom\\
    \textsuperscript{3}School of Advanced Technology, Xi'an Jiaotong-Liverpool University, China\\
    \textsuperscript{4}School of Physical Sciences, University of Liverpool, Liverpool, United Kingdom\\
    \textsuperscript{5}School of Mathematics and Physics, Xi'an Jiaotong-Liverpool University, China\\
    \textsuperscript{6}Institute of Artificial Intelligence Innovation, Zhejiang University of Technology, China\\
    \textsuperscript{7}Urban Governance and Design Thrust, The Hong Kong University of Science and Technology (Guangzhou), China}

}

\maketitle
\begin{abstract}
A structured understanding of intraoral 3D scans is essential for digital orthodontics. However, existing deep-learning approaches rely heavily on modality-specific training, large annotated datasets, and controlled scanning conditions, which limit generalization across devices and hinder deployment in real clinical workflows. Moreover, raw intraoral meshes exhibit substantial variation in arch pose, incomplete geometry caused by occlusion or tooth contact, and a lack of texture cues, making unified semantic interpretation highly challenging. To address these limitations, we propose ArchMap, a training-free and knowledge-guided framework for robust structured dental understanding. ArchMap first introduces a geometry-aware arch-flattening module that standardizes raw 3D meshes into spatially aligned, continuity-preserving multi-view projections. We then construct a Dental Knowledge Base (DKB) encoding hierarchical tooth ontology, dentition-stage policies, and clinical semantics to constrain the symbolic reasoning space. Building on this ontology, a schema-constrained vision--language inference pipeline transforms general-purpose VLMs into deterministic, contract-compliant structured predictors. We validate ArchMap on 1060 pre-/post-orthodontic cases, demonstrating robust performance in tooth counting, anatomical partitioning, dentition-stage classification, and the identification of clinical conditions such as crowding, missing teeth, prosthetics, and caries. Compared with supervised pipelines and prompted VLM baselines, ArchMap achieves higher accuracy, reduced semantic drift, and superior stability under sparse or artifact-prone conditions. As a fully training-free system, ArchMap demonstrates that combining geometric normalization with ontology-guided multimodal reasoning offers a practical and scalable solution for the structured analysis of 3D intraoral scans in modern digital orthodontics.
\end{abstract}

\begin{IEEEkeywords}
Tooth Counting, Dental Understanding, Vision-language Model
\end{IEEEkeywords}

\section{Introduction}

With the widespread adoption of dental digitization devices and workflows, digital dentistry is reshaping the end-to-end orthodontic pipeline: from efficiently acquiring three-dimensional morphology of the maxillary and mandibular dental arches using intraoral scanners (IOS), to digital analysis and automated assessment during treatment planning and follow-up ~\cite{liu2023deep,chen2024deep,alassiry2023clinical}. Compared with two-dimensional imaging, IOS data provide high-resolution arch geometry without radiation, enabling tooth identification, surface analysis, occlusal relationship evaluation, and outcome quantification; consequently, an increasing number of clinical and research tasks are conducted on 3D meshes (this study likewise uses a deidentified IOS dataset primarily consisting of STL arch meshes) ~\cite{wu2022two,zanjani2021mask,im2022accuracy,wu2024transformer,zhang2020automatic}. However, extracting task-relevant structured semantics from raw IOS data efficiently and robustly, while maintaining clinical interpretability, remains a challenging research problem ~\cite{kot2025evolution,yang2025clinica}.

Despite recent progress in 3D understanding for dental applications—such as tooth segmentation and enumeration on point clouds/meshes, multi-view rendering-based recognition and measurement ~\cite{miao2025dentalsplat}, and multi-scale geometric modeling that combines full-arch context with local patches—substantial challenges persist: large variations in dental arch pose; geometry missing due to occlusion and contact; confusion caused by morphological similarity among teeth; high annotation cost with inconsistent labeling styles; and limited generalization across devices and populations ~\cite{nambiar2024comprehensive,wang2025application}. Existing methods typically struggle to balance abundant global context with sufficient local geometric detail, leading to trade-offs when modeling fine-grained targets (e.g., localized surface defects, mild rotations/crowding) versus global relationships (e.g., maxilla-mandible registration, quadrant symmetries) ~\cite{xi20253d,chen2025cross,lian2020deep,Chen2021Hierarchical,hao2022toward}.

Furthermore, IOS data are widely stored in STL format and thus lack color/texture cues, making it difficult to leverage appearance differences. They also lack explicit semantics (e.g., tooth indices, anatomical parts, and surface labels), so downstream tasks—such as FDI-based tooth indexing, cross-subject statistics, and longitudinal comparisons—depend on additional enumeration and alignment steps. Artifacts and holes from the scanning process, inconsistent coordinate frames and scales, and the under-utilization of symmetry and semantic constraints between the maxillary/mandibular arches and across quadrants further limit joint modeling of fine-grained details and full-arch structure ~\cite{Wang2020Multiview,Lian2021MeshSegNet}.

To address these challenges, we propose \textbf{ArchMap}, a geometry-aware and arch-preserving multi-view mapping framework for structured dental understanding from intraoral scan models. ArchMap performs standardized \emph{arch flattening} along the dental curve to convert 3D meshes into spatially aligned multi-view 2D representations, preserving anatomical continuity while enhancing visual interpretability. It further introduces a \emph{task-specific Dental Knowledge Base (DKB)} that encodes hierarchical dental ontology, serving as symbolic constraints and task priors. Finally, a \emph{training-free vision--language inference mechanism} is applied to perform structured reasoning over the generated views, enabling unified outputs compatible with clinical workflows such as tooth counting, arch partitioning, and dentition-stage assessment.

The main contributions of this work are summarized as follows:
\begin{itemize}
  \item We propose \textbf{ArchMap}, a geometry-aware arch flattening strategy that standardizes 3D dental meshes into multi-view 2D projections while preserving arch continuity and spatial alignment, thereby enabling accurate vision--language inference without additional training.  
  \item We design a task-specific \textbf{Dental Knowledge Base (DKB)} that encodes FDI tooth indexing, dentition stages, and clinical conditions in a hierarchical ontology. DKB provides symbolic guidance that reduces semantic ambiguity and enhances the robustness of structured dental reasoning.  
  \item We develop a zero-training, schema-constrained vision--language inference framework that treats VLMs as structured reasoning engines. The integration of geometric flattening, ontology knowledge, and DKB guidance ensures unified, interpretable, and clinically consistent outputs, as demonstrated by comparative and ablation experiments.  
\end{itemize}

\section{Related Work}

\subsubsection{Deep Learning for 3D Tooth Counting}
Early approaches to automatic tooth counting in 3D data relied on geometric reasoning to segment individual teeth. These methods exploited features like surface curvature or active contours to separate teeth from gingiva, but they often lacked robustness to anatomical variations and required manual intervention for initialization ~\cite{beser2024yolo,ong2024fully}. In recent years, supervised deep learning has substantially advanced 3D tooth segmentation and enumeration. For example, Jang \textit{et al.} developed a hierarchical multi-step CNN for CBCT volumes: their system projects the 3D scan into panoramic images to detect each tooth, then refines segmentation within 3D regions-of-interest ~\cite{jang2021fully}. This approach achieved an F1-score of 93.3\% for tooth identification and a Dice coefficient of 94.8\% for individual tooth segmentation on CBCT data, demonstrating near-expert accuracy. Similarly, Cui \textit{et al.} presented a large-scale AI model that segments teeth from CBCT scans with an average Dice score of ~91.5\%, comparable to experienced radiologists ~\cite{cui2022fully}. For surface scans (STL/IOS), researchers have explored both 2D and 3D strategies: some convert 3D dental meshes into 2D images for CNN processing, while others apply point cloud networks directly ~\cite{zhang2020automatic,wu2024transformer}. These techniques can enumerate full dentitions in digital models, though each modality (CBCT vs. IOS) typically demands separate training and carefully curated annotations. Notably, weakly-supervised frameworks are emerging to reduce annotation burden – e.g., the recent SAMTooth method leverages the Segment-Anything Model (SAM) with one-point prompts per tooth to guide 3D point cloud segmentation, reaching accuracy on par with fully-supervised methods using only 0.1\% labeled points ~\cite{liu20243d}.

\subsubsection{Vision-Language Models in Dental Applications}
With the rise of large-scale vision-language models (VLMs) in healthcare, the dental domain has started to explore their potential. The primary example is DentVLM by Meng \textit{et al.}, a multimodal model trained on 110,447 images with 2.46 million paired question-answers covering a wide range of oral health topics ~\cite{meng2025dentvlm}. DentVLM is capable of interpreting seven distinct 2D dental imaging modalities and performing 36 diagnostic tasks, significantly outperforming previous models (by 19.6\% in disease identification accuracy). It achieved near-expert results on many benchmarks – even surpassing junior dentists and matching some senior experts on certain diagnostic tasks – illustrating the power of foundation models for dental intelligence. Beyond DentVLM’s comprehensive scope, other works have targeted specific vision-language challenges. Du \textit{et al.} employed a pre-trained detector (Grounding DINO) with tailored text prompts to perform \textit{dental notation-aware} detection on panoramic X-rays, effectively enumerating teeth and pinpointing abnormalities by leveraging the symmetric oral anatomy and standard tooth codes ~\cite{du2024prompting}. In the realm of purely language-based models, domain-specific LLMs such as \textit{DentalGPT} have been introduced, fine-tuned on dental texts to assist in patient Q\&A and clinical decision support; however, these text-only models inherently struggle with visual-spatial tasks ~\cite{hou2025benchmarking}. On the vision side, foundation segmentation models have also been adapted for dentistry: for instance, DentalSAM refers to leveraging Meta’s Segment-Anything Model for dental images, enabling promptable segmentation of teeth with minimal user input ~\cite{li2024adapting}. Such prompt-driven approaches have even been extended to 3D data, as noted above with SAMTooth ~\cite{liu20243d}. Another notable effort is the MMOral benchmark and OralGPT model by Zhang \textit{et al.} ~\cite{hao2025towards}, which evaluated dozens of general-purpose LVLMs on dental radiograph interpretation. Their findings revealed that off-the-shelf models (e.g., GPT-4V) barely reach 41\% accuracy on domain-specific questions, whereas a custom-tuned model (OralGPT, fine-tuned on 1.3 million dental instructions) achieved a dramatic 24.7\% performance gain. This underscores the importance of adapting large VLMs to dental data and tasks.

\section{Methodology}
\subsection{Geometry-Aware Dental Arch Flattening}

To transform irregular 3D intraoral scan meshes into anatomically consistent 2D representations, \textbf{ArchMap} employs a geometry-aware dental arch flattening strategy that explicitly models arch curvature and occlusal topology. The process consists of three stages: (1) parabola-based arch estimation via rotational regression; (2) normal-direction surface flattening; and (3) continuity-preserving rendering for multi-view generation.

\subsubsection{Parabola-Based Arch Estimation}

Given a triangle mesh \( M = (V, F) \) with vertex set \(V \in \mathbb{R}^{N \times 3}\), the 2D occlusal projection is first centralized using its centroid \((\bar{X}, \bar{Y})\). A rotational grid search is then performed within the range of \([-90^{\circ}, 90^{\circ}]\). For each candidate rotation angle \(\theta\), a parabola is fitted in the rotated coordinate system as:

\begin{equation}
y' = a x'^2 + b x' + c.
\end{equation}
The optimal rotation angle \(\theta^*\) and coefficients \((a,b,c)\) are determined by minimizing the residual error:
\begin{equation}
\min_{a,b,c,\theta} \sum_i \big( y'_i - (a x'^2_i + b x'_i + c) \big)^2.
\end{equation}
A hierarchical two-stage (coarse-to-fine) grid search ensures stable optimization under different orientations. To enhance robustness, extreme residuals are clipped at the 95th percentile, resulting in a reliable global dental arch curve~\cite{ren2024high}.

\begin{figure*}[htbp]
  \centering
    \includegraphics[width=\linewidth]{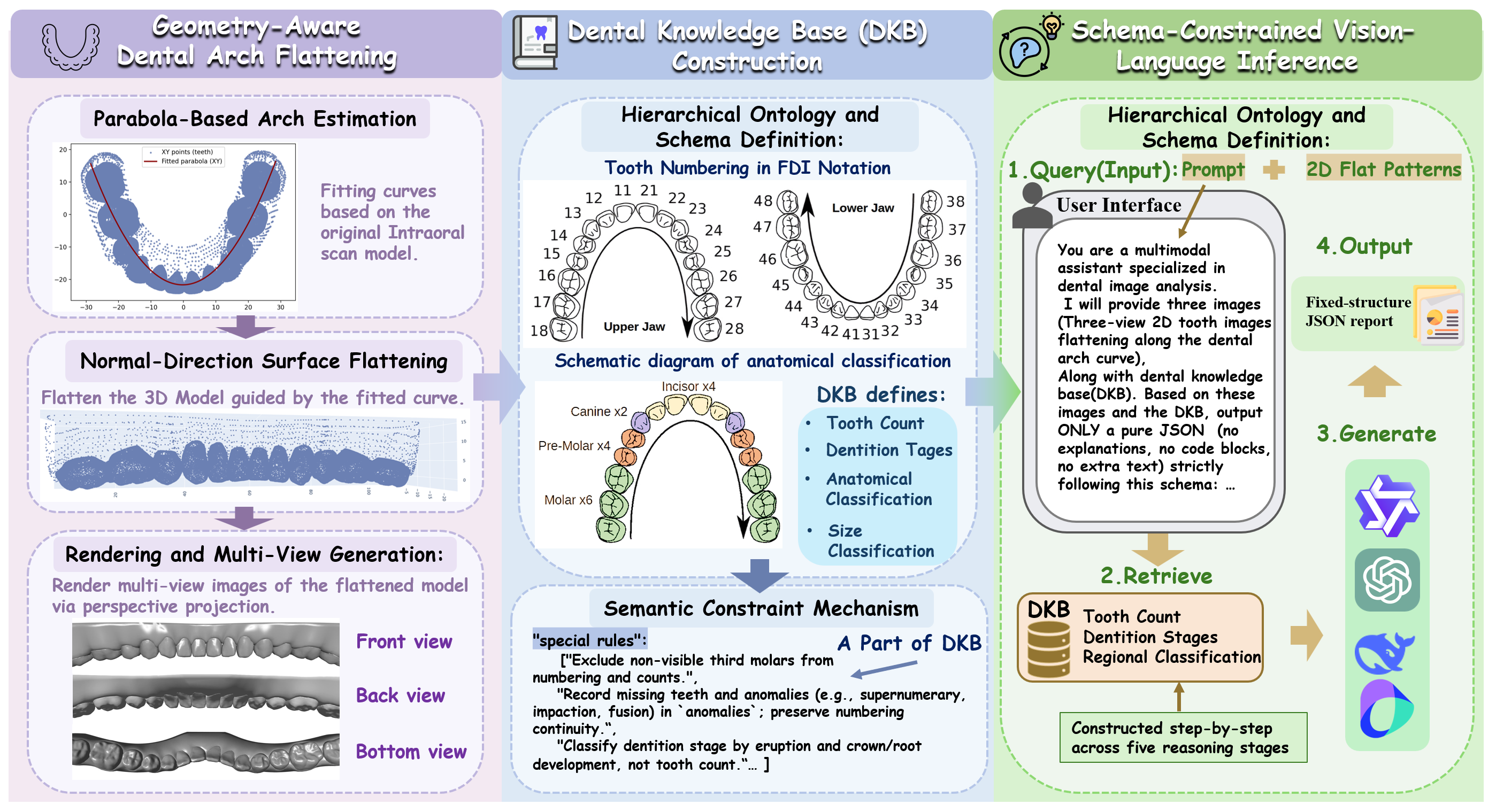}
    \vspace{-1.5em}
  \caption{\textbf{ArchMap workflow (left to right).} 
\textbf{Left:} Geometry-aware dental arch flattening—parabola-based curve estimation, normal-direction surface flattening, and continuity-preserving rendering to obtain standardized front/back/bottom views. 
\textbf{Middle:} Dental Knowledge Base (DKB)—a lightweight hierarchical ontology defining tooth count, dentition stages, regional and size categories, and semantic constraints. 
\textbf{Right:} Schema-constrained vision–language inference—a four-stage procedure (Query → Retrieve → Generate → Output) executed on a frozen VLM to produce a fixed-structure JSON report.}
\label{fig:workflow}

\end{figure*}

\subsubsection{Normal-Direction Surface Flattening}

After obtaining the reference curve, the parabola is uniformly sampled along the arc length \(s\) to maintain spatial consistency:
\begin{equation}
s_i = \int_{x_0}^{x_i} \sqrt{1+\left(\frac{dy}{dx}\right)^2} \, dx.
\end{equation}
Each vertex \(v=(x',y',z)\) is projected to its nearest point \((x_c, y_c)\) on the fitted curve, and the orthogonal offset along the local normal direction \(\vec{n}=(-\tfrac{dy}{dx}, 1)\) is computed as:
\begin{equation}
d_n = (x' - x_c)n_x + (y' - y_c)n_y.
\end{equation}
The flattened coordinates are then defined as:
\begin{equation}
v_{\text{flat}} = (s_c,\, d_n + y_v,\, z).
\end{equation}
This transformation straightens the curved dental arch into a canonical 2D layout while preserving occlusal geometry and vertical consistency.

\subsubsection{Rendering and Multi-View Generation}

To facilitate visual interpretation and downstream vision--language reasoning, the flattened mesh is rendered with physically inspired metallic shading using \textit{Plotly--Kaleido}~\cite{xi20253d}. Standardized front, back, and bottom views are captured and cropped to a consistent aspect ratio, forming an ordered multi-view set \(I_{\text{arch}} \in \mathbb{R}^{n \times H \times W \times 3}\)~\cite{wang2025structure}. This representation preserves both anatomical continuity and spatial topology, providing geometry-aligned 2D inputs for subsequent structured reasoning.

\subsection{Dental Knowledge Base (DKB) Construction}

To achieve ontology-aligned and clinically interpretable reasoning, ArchMap introduces a task-specific \textbf{Dental Knowledge Base (DKB)}.  
The DKB defines hierarchical relations of dental anatomy, developmental stages, and clinical states, providing symbolic priors for vision--language reasoning.  
As a lightweight schema, it unifies output structure across cases and enhances semantic stability and interpretability~\cite{debellis2024integrating,schleyer2013ontology}.

\subsubsection{Hierarchical Ontology and Schema Definition}

The DKB is organized around four core semantic domains: \textit{tooth count}, \textit{dentition stage}, \textit{regional classification}, and \textit{semantic constraint mechanism}.  
Its goal is to establish structured mappings between visual features and clinical semantics.

\paragraph{Tooth Count}
The DKB encodes canonical definitions for both deciduous and permanent dentitions.  
Deciduous dentition typically consists of 20 teeth (5 per quadrant), while permanent dentition includes 28--32 teeth depending on the eruption of third molars.  
This representation specifies ideal per-arch and per-quadrant distributions while allowing flexibility for clinical variations such as extractions or congenital absence.

\paragraph{Dentition Stages}
Three developmental stages are defined: \textit{deciduous}, \textit{mixed}, and \textit{permanent}.  
These stages are determined not only by chronological factors but also by eruption patterns and morphological features.  
The deciduous stage corresponds to infancy and early childhood, the mixed stage represents the transitional coexistence of primary and permanent teeth, and the permanent stage is established after approximately twelve years of age.

\paragraph{Regional Classification}
In anatomical terms, the DKB divides the dentition into three functional regions: anterior, premolar, and molar, responsible respectively for cutting, transitional chewing, and grinding.  
Additionally, the DKB defines a morphological hierarchy where \textit{Large} corresponds to molars, \textit{Medium} to canines, premolars, and central incisors, and \textit{Small} to lateral incisors, achieving consistent encoding across shape, function, and geometry.

\subsubsection{Semantic Constraint Mechanism}

To ensure anatomical plausibility and clinical validity, the DKB embeds a set of formalized semantic constraints that delimit the model's reasoning space.  
The valid inference domain is defined as:
\begin{equation}
\mathcal{I} = 
\Bigl\{
t_i \in \mathcal{T}_v \ \Big|\ 
t_i \in \mathcal{A},\ 
\text{state}(t_i) \in \mathcal{S},\
\text{semantic constraints}
\Bigr\},
\end{equation}
where $\mathcal{A}$ denotes the set of valid dental arches, $\mathcal{T}_v$ the set of visible teeth, and $\mathcal{S}$ the space of admissible clinical states (e.g., missing, restored, or carious).

The following semantic constraints further regulate inference consistency and clinical validity:
\begin{itemize}
    \item \textbf{Conditional inclusion:} third molars are considered only when supported by visual evidence.
    \item \textbf{Numbering continuity:} missing, extracted, or supernumerary teeth must be explicitly documented to preserve index order.
    \item \textbf{Morphology-guided classification:} dentition-stage assessment is constrained by morphological and eruption-based criteria rather than numerical count alone.
    \item \textbf{Positional annotation:} orthodontic extractions must retain precise arch-level and spatial information.
\end{itemize}

Together, these constraints define the model’s \textit{semantic feasible domain}, ensuring that all reasoning outcomes remain anatomically coherent, numerically consistent, and clinically interpretable under zero-training conditions.

\vspace{-1em}
\subsection{Schema-Constrained Vision--Language Inference}

Building upon the semantic constraint mechanism, ArchMap introduces a \textit{schema-constrained vision--language inference framework} to achieve structured dental understanding under zero-training conditions.  
This framework reformulates a vision--language model (VLM) from a free-form generator into a \textit{structured reasoning engine}, ensuring that its outputs strictly adhere to the symbolic ontology and schema defined by the Dental Knowledge Base (DKB)~\cite{ayadiontology,du2024prompting,fanelli2025development}.

Given three multi-view 2D projections of a single dental arch 
\(\{I_\text{front}, I_\text{back}, I_\text{bottom}\}\) 
and the knowledge schema \(\mathcal{K}\), the inference process is formally expressed as:
\begin{equation}
\hat{Y} = f_\theta \big( I_\text{front}, I_\text{back}, I_\text{bottom}; \mathcal{K} \big),
\quad \text{s.t. } \hat{Y} \in \mathcal{K},
\end{equation}
where \(f_\theta\) denotes the frozen multimodal backbone of the VLM, and \(\mathcal{K}\) defines the valid symbolic space of dental attributes, including anatomical regions, size hierarchies, dentition stages, and clinical states.  
Through this constraint, all generated outputs are aligned with the DKB ontology, guaranteeing determinism, structural consistency, and clinical interpretability.

\vspace{-0.5em}

The inference process in ArchMap unfolds through five hierarchical reasoning stages that progressively construct a clinically interpretable representation of the dental arch.  
It begins with \textbf{tooth counting}, in which visible crowns are identified and the total number of teeth within the arch is estimated.  
Subsequently, \textbf{anatomical classification} assigns each detected tooth to one of three functional regions: anterior, premolar, or molar, according to its spatial position and morphology.  
In the next \textbf{size classification} stage, each tooth is categorized as large, medium, or small based on the hierarchical definitions encoded in the DKB.  
Then, \textbf{dentition stage determination} integrates eruption patterns and morphological cues to infer whether the arch corresponds to a deciduous, mixed, or permanent dentition.  
Finally, \textbf{clinical condition identification} detects abnormalities such as missing teeth, prosthetics, or caries and records them as symbolic annotations in a structured format. Fig.~\ref{fig:workflow} illustrates the workflow of this paper, with the three components of the Methodology represented from left to right.

\section{Experiments and Results}

\begin{table*}[!t]
\centering
\begin{threeparttable}
\caption{Unified Results with FDI Taxonomy and Dentition-Stage Policies.}
\label{tab:unified-big}
\renewcommand{\arraystretch}{1.05}
\setlength{\tabcolsep}{4pt}
\begin{tabular}{l l p{0.24\textwidth} c c c c c}
\toprule
\textbf{Task} & \textbf{Category} & \textbf{Definition / Notes\tnote{2}} & \textbf{Typical Count} & \textbf{Pred} & \textbf{AE} & \textbf{RE(\%)} & \textbf{Acc(\%)} \\
\midrule
\multirow{2}{*}{Tooth Counting}
& Arch (Upper) & --- & --- & 12.034 & 3.987 & 24.886 & 75.114 \\
& Arch (Lower) & --- & --- & 12.176 & 3.912 & 24.316 & 75.684 \\
\cmidrule(lr){1-8}
\multirow{3}{*}{Anatomical Classification}
& Anterior
& FDI ranges (Q1–Q4): 11–13; 21–23; 31–33; 41–43
& 12 & 9.577 & 2.423 & 20.192 & 79.808 \\
& Premolars
& 14–15; 24–25; 34–35; 44–45
& 8 & 6.216 & 1.784 & 22.300 & 77.700 \\
& Molars
& 16–18; 26–28; 36–38; 46–48
& 12 & 9.088 & 2.912 & 24.267 & 75.733 \\
\cmidrule(lr){1-8}
\multirow{3}{*}{Size Classification}
& Large
& Molars: 16–18; 26–28; 36–38; 46–48
& 12 & 9.403 & 2.597 & 21.642 & 78.358 \\
& Medium
& Canines (13, 23, 33, 43); Premolars (14–15, 24–25, 34–35, 44–45); Centrals (11, 21, 31, 41)
& 16 & 12.787 & 3.213 & 20.081 & 79.919 \\
& Small
& Lateral incisors: 12, 22, 32, 42
& 4 & 3.098 & 0.902 & 22.550 & 77.450 \\
\cmidrule(lr){1-8}
\multirow{3}{*}{Dentition Stage\tnote{1}}
& Primary
& Exclude third molars; anomalies logged in \texttt{special\_conditions}.
& 20 & 18.963 & 1.037 & 5.185 & 94.815 \\
& Mixed
& Transitional stage; no auto-completion.
& variable & 21.947 & 7.036 & 24.276 & 75.724 \\
& Permanent
& Explicitly state third-molar inclusion.
& 28--32 & 22.043 & 5.957 & 21.275 & 78.725 \\
\bottomrule
\end{tabular}
\begin{tablenotes}
\footnotesize
\item[1] Stage rows report \emph{stage accuracy} in the Acc column; other metrics are not applicable for stage evaluation and are reported only for completeness.
\item[2] For Anatomical Position and Tooth Size, this column lists FDI indices in compact ranges; inclusion of third molars must be explicitly stated in the protocol.
\end{tablenotes}
\end{threeparttable}
\end{table*}

\subsection{Data Collection and Hyperparameter Settings}

We adopt the public dataset A 3D Dental Model Dataset~\cite{wang20243d}with Pre/Post-Orthodontic Treatment (v4), containing 1,060 pre/post pairs from 435 subjects. Raw STL meshes are converted into multi-view 2D projections via geometry-aware flattening (see \S2.1.1). Key settings are as follows.

\paragraph*{Arch fitting}
The dental arch is fitted by angular sampling around the model centroid with a coarse step of $1.0^{\circ}$ and local refinement within a $5.0^{\circ}$ half-window for up to two iterations. 
Points beyond the $0.95$ quantile of radial distances are removed via robust clipping, and spatial neighborhoods are efficiently retrieved using the \texttt{cKDTree} structure~\cite{narasimhulu2021ckd} for fast nearest-neighbor searches.

\paragraph*{Rendering and camera configuration}
Each view is defined by an extrinsic transformation $\mathbf{T}=[\mathbf{R}|\mathbf{t}]$, 
where $\mathbf{t}$ denotes the camera center and $\mathbf{R}$ orients the optical axis toward the model centroid. 
For the maxillary arch, three canonical viewpoints are adopted:
\begin{equation}
\mathbf{t}_{\text{front}}{=}(0,2.6,0),~
\mathbf{t}_{\text{back}}{=}(0,-2.6,0),~
\mathbf{t}_{\text{bottom}}{=}(0,0,-2.6).
\end{equation}

For the mandibular arch, the bottom view is mirrored along the $z$-axis:
\begin{equation}
\mathbf{t}_{\text{bottom}} = (0,\,0,\,2.6).
\end{equation}
All cameras share a common up vector $\mathbf{u}=(1,\,0,\,0)$ and are placed at a normalized distance of $2.6$ units from the model centroid, corresponding to an effective focal length of $f=35\,\mathrm{mm}$ in normalized model units. 
The optical axes of all cameras point toward the origin, ensuring geometric consistency across renders. 
This formulation allows reproducible multi-view projection for both maxillary and mandibular models under identical camera intrinsics and controlled viewpoints.

\begin{table*}[h]
  \centering
  \begin{threeparttable}
  \caption{Overall performance across VLMs with thinking vs. non-thinking modes, with appended prior methods (percentages are in \%).}
  \label{tab:vlm_all_with_modes_refined_combined}
  \setlength{\tabcolsep}{1.2pt}
  \renewcommand{\arraystretch}{1.15}
  \footnotesize

  \newcommand{\up}{\textcolor{green!60!black}{\small$\uparrow$}}
  \newcommand{\down}{\textcolor{red!70!black}{\small$\downarrow$}}

  \newcommand{\gcell}[1]{\cellcolor{gray!5}{#1}}      
  \newcommand{\gcellb}[1]{\cellcolor{gray!5}{\textbf{#1}}} 
  \newcommand{\Gcell}[1]{\cellcolor{gray!12}{#1}}     
  \newcommand{\Gcellb}[1]{\cellcolor{gray!12}{\textbf{#1}}}

  \begin{tabular}{l l cccccccccccc}
    \toprule
    \multirow{3}{*}{\textbf{Method}} & \multirow{3}{*}{\textbf{Category(Mode)}}
      & \multicolumn{5}{c}{\textbf{Counting}} & \multicolumn{7}{c}{\textbf{Partition (F1)}} \\
    \cmidrule(lr){3-7}\cmidrule(lr){8-14}
     &  & AE~\down & RE~\down & Acc~\up & Over~\down & Under~\down & Macro\mbox{-}F1~\up
     & \multicolumn{3}{c}{Maxillary~(\up)} & \multicolumn{3}{c}{Mandibular~(\up)} \\
    \cmidrule(lr){9-11}\cmidrule(lr){12-14}
     &  &  &  &  &  &  &  & Ant & Pre & Mol & Ant & Pre & Mol \\
    \midrule

    \multicolumn{14}{l}{\textbf{Prior Methods }} \\
    \addlinespace[2pt]

    \multicolumn{1}{l}{YOLOv5~\cite{beser2024yolo}}
    & Supervised Det./Seg.
    & \gcellb{8.954} & \gcellb{27.981} & \gcellb{72.019} & \gcellb{14.665} & \gcellb{13.316} & \gcellb{71.582}
    & \gcellb{72.429} & \gcellb{71.787} & \gcellb{70.883} & \gcellb{71.983} & \gcellb{71.581} & \gcellb{70.829} \\

    \multicolumn{1}{l}{Demirjian staging pipeline~\cite{ong2024fully}}
    & Deep Stage Class.
    & 11.514 & 35.983 & 64.017 & 18.859 & 17.124 & 63.422
    & 64.819 & 63.711 & 62.471 & 64.171 & 63.377 & 61.983 \\

    \multicolumn{1}{l}{Prompted VLM~\cite{du2024prompting}}
    & VLM-based Reasoning
    & 9.756 & 30.481 & 69.519 & 15.975 & 14.506 & 69.133
    & 70.117 & 69.383 & 68.329 & 69.823 & 69.129 & 68.017 \\

    \addlinespace[4pt]
    \midrule

    \multicolumn{14}{l}{\textbf{VLMs (Model in col-1, Mode in col-2)}} \\
    \addlinespace[2pt]

    \rowcolor{gray!5}
    \multirow{2}{*}{\cellcolor{white}Qwen3\mbox{-}VL\mbox{-}235B\mbox{-}A22B\cite{cai2025comparebench}\mbox
    }
      & Thinking
        & \textbf{3.486} & \textbf{21.863} & \textbf{78.137} & \textbf{11.181} & \textbf{10.682} & \textbf{78.237}
        & \textbf{80.083} & \textbf{78.613} & \textbf{76.914} & \textbf{79.321} & \textbf{77.832} & \textbf{76.487} \\
      & Non\mbox{-}Thinking
        & 3.721 & 25.121 & 74.879 & 12.847 & 12.274 & 77.103
        & 79.214 & 77.823 & 76.073 & 78.482 & 76.942 & 75.721 \\

    \addlinespace[2pt]

    \rowcolor{gray!5}
    \multirow{2}{*}{\cellcolor{white}Qwen3\mbox{-}VL\mbox{-}30B\mbox{-}A3B\cite{xu2025qwen3}\mbox}
      & Thinking
        & \textbf{3.698} & \textbf{22.517} & \textbf{77.483} & \textbf{11.809} & \textbf{10.708} & \textbf{77.463}
        & \textbf{79.176} & \textbf{77.892} & \textbf{75.824} & \textbf{78.362} & \textbf{76.914} & \textbf{76.821} \\
      & Non\mbox{-}Thinking
        & 3.926 & 25.345 & 74.655 & 13.292 & 12.053 & 76.379
        & 78.311 & 77.051 & 75.012 & 77.522 & 76.012 & 75.983 \\

    \addlinespace[2pt]

    \rowcolor{gray!5}
    \multirow{2}{*}{\cellcolor{white}GLM\mbox{-}4.5V\cite{zhu2025assessing}}
      & Thinking
        & \textbf{3.824} & \textbf{22.974} & \textbf{77.026} & \textbf{12.086} & \textbf{10.888} & \textbf{76.789}
        & \textbf{78.614} & \textbf{76.782} & \textbf{75.097} & \textbf{77.583} & \textbf{75.887} & \textbf{76.832} \\
      & Non\mbox{-}Thinking
        & 4.051 & 26.748 & 73.252 & 14.072 & 12.676 & 75.684
        & 77.742 & 75.972 & 74.264 & 76.712 & 75.071 & 75.937 \\

    \addlinespace[2pt]

    \rowcolor{gray!5}
    \multirow{2}{*}{\cellcolor{white}doubao\mbox{-}seed\mbox{-}1\mbox{-}6 (vision)\cite{huang2025memorb}
}
      & Thinking
        & \textbf{3.917} & \textbf{23.356} & \textbf{76.644} & \textbf{12.476} & \textbf{10.880} & \textbf{76.243}
        & \textbf{78.012} & \textbf{76.214} & \textbf{74.611} & \textbf{77.108} & \textbf{75.327} & \textbf{76.018} \\
      & Non\mbox{-}Thinking
        & 4.168 & 26.459 & 73.541 & 14.134 & 12.325 & 75.042
        & 77.162 & 75.341 & 73.741 & 76.252 & 74.474 & 75.141 \\

    \addlinespace[4pt]
    \midrule

    \multicolumn{14}{l}{\textbf{Baseline}} \\
    \addlinespace[2pt]

    \rowcolor{gray!12}
    \multirow{2}{*}{\cellcolor{white}\textbf{VLM Aggregate (4 models)}}
      & \textbf{Thinking}
        & \textbf{3.950} & \textbf{24.601} & \textbf{75.399} & \textbf{12.892} & \textbf{11.709} & \textbf{77.183}
        & \textbf{78.971} & \textbf{77.375} & \textbf{75.612} & \textbf{78.094} & \textbf{76.490} & \textbf{76.540} \\
      & \Gcell{\textbf{Non\mbox{-}Thinking}}
        & \Gcellb{\textbf{3.966}} & \Gcellb{\textbf{25.918}} & \Gcellb{\textbf{74.082}} & \Gcellb{\textbf{13.582}} & \Gcellb{\textbf{12.336}} & \Gcellb{\textbf{76.052}}
        & \Gcellb{\textbf{78.107}} & \Gcellb{\textbf{76.547}} & \Gcellb{\textbf{74.772}} & \Gcellb{\textbf{77.242}} & \Gcellb{\textbf{75.625}} & \Gcellb{\textbf{75.696}} \\

    \bottomrule
  \end{tabular}
  \end{threeparttable}
\end{table*}

\subsection{Quantitative Evaluation}
\label{sec:quant}

\subsubsection{Unified Taxonomy Evaluation: Counting, Anatomy, and Size}

We evaluate three objectives on the unified taxonomy dataset: (i) full-arch tooth counting, (ii) anatomical region classification (anterior, premolar, molar), and (iii) size-based categorization (large, medium, small). Results are computed per arch (maxillary/mandibular) and aggregated to the case level. Unless otherwise specified, only JSON-compliant outputs are included to ensure consistency with the FDI numbering and dentition-stage protocol. Table~\ref{tab:unified-big} reports the consolidated statistics.

For numerical predictions we report absolute error (AE), relative error (RE), and accuracy (Acc), all reported in percentages (\%):
\begin{equation*}
\mathrm{AE}=|\hat{y}-y|,\;
\mathrm{RE}=\tfrac{|\hat{y}-y|}{y},\;
\mathrm{Acc}=\max\!\left(0,\,1-\tfrac{|\hat{y}-y|}{y}\right).
\end{equation*}
Over/Under counting is summarized as the signed deviation $\hat{y}-y$.
For categorical partitions (anatomy/size), we use macro-averaged F1:
\begin{equation*}
\mathrm{F1}_{\mathrm{macro}}
=\tfrac{1}{|\mathcal{C}|}\!\sum_{c\in\mathcal{C}}\!
\tfrac{2P_cR_c}{P_c+R_c},\;
P_c=\tfrac{TP_c}{TP_c+FP_c},\;
R_c=\tfrac{TP_c}{TP_c+FN_c}.
\end{equation*}
Dentition-stage accuracy is
$\mathrm{StageAcc}=\tfrac{1}{N}\sum_i \mathbb{1}(\hat{s}_i=s_i)$.

When only per-tooth statistics are available, detection counts are mapped to counting metrics as:
\begin{equation*}
\begin{aligned}
\mathrm{Pred}&=TP{+}FP, \qquad & \mathrm{GT}&=TP{+}FN,\\
\mathrm{AE}&=|FP{-}FN|, \qquad & 
\mathrm{RE}&=\tfrac{|FP{-}FN|}{TP{+}FN}.
\end{aligned}
\end{equation*}

Under the unified protocol (Table~\ref{tab:unified-big}), counting shows similar behavior on both arches with RE $\approx\!24\%$ (Acc $\approx\!75\%$). In anatomical partitions, anterior teeth are most stable (Acc $\sim\!80\%$) owing to regular morphology and clear visibility; molars are more challenging ($\sim\!76\%$) due to occlusal complexity, metallic restorations, and soft-tissue occlusions. In size-based partitions, medium classes (centrals, canines, premolars) yield the most balanced performance (Acc $\approx\!79.9\%$), while small laterals may be under-recalled in crowded contacts and large molars are occasionally affected by glare. Stage recognition remains high for primary ($\sim\!94.8\%$) and stable for permanent ($\sim\!78.7\%$), indicating robustness across growth phases. Overall, errors concentrate in anatomically ambiguous or visibility-limited regions rather than being uniformly scattered.

\vspace{0.5em}

\subsubsection{Cross-Method and Cross-Mode Comparison}

We benchmark the proposed framework against reproduced supervised pipelines (detection/segmentation and staging) and a prompted VLM baseline under identical decoding and formatting controls (temperature~0, strict JSON schema, unified post-processing). This isolates the contribution of our method from decoding effects.

Table~\ref{tab:vlm_all_with_modes_refined_combined} shows that our approach consistently reduces counting error, improves arch-level accuracy, and raises partition F1 across both jaws, while lowering both over- and under-count rates. Supervised pipelines exhibit boundary brittleness (premolar–molar transitions; reflective/occluded posterior regions). The prompted VLM baseline, though stronger than classic detectors on taxonomy, remains sensitive to posterior visibility and tends to drift around mandibular molars. Qualitative checks mirror these trends: supervised methods over-detect near reflective restorations, whereas the prompted baseline under-recalls small laterals in crowded contacts.

Across four representative VLM backbones, \emph{Thinking} consistently outperforms \emph{Non-Thinking} under identical settings: gains are uniform at the arch level and most pronounced in molar partitions (the principal visibility bottleneck). Aggregated results show higher accuracy and macro-F1, reduced over/under bias, and tighter cross-model variance, indicating improved stability rather than backbone-specific effects. We also observe fewer schema deviations under strict JSON decoding, suggesting better adherence to the output contract and more reliable multi-view evidence integration. Overall, \emph{Thinking} is the preferred inference configuration across heterogeneous VLM.

\subsection{Special Clinical Conditions}

\begin{figure*}[!t]
  \centering
    \includegraphics[width=\linewidth]{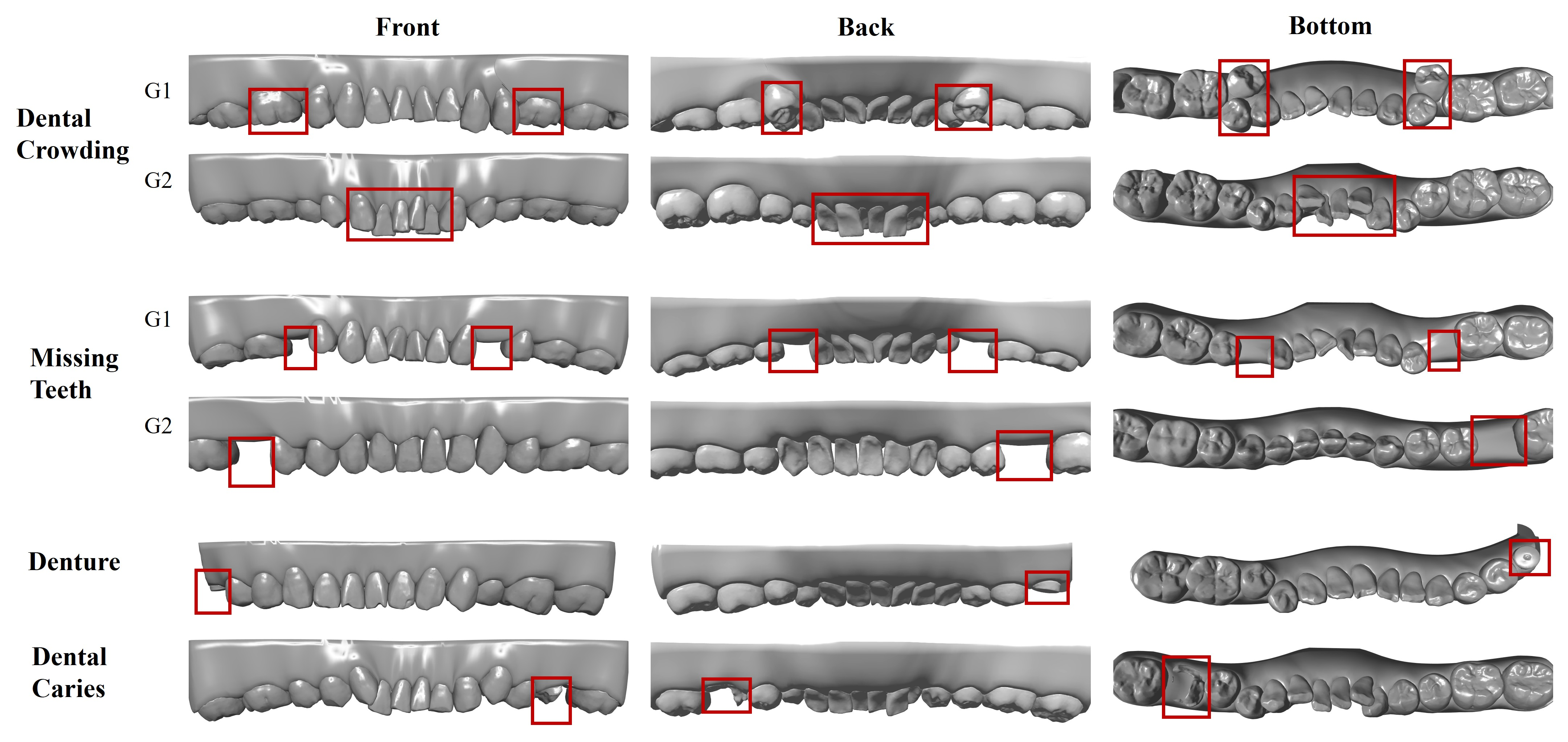}
  \caption{Qualitative observations of four representative clinical conditions: \emph{Dental Crowding}, \emph{Missing Teeth}, \emph{Denture}, and \emph{Dental Caries}. Each case is shown from the front, back, and bottom views; red boxes mark regions attended by the model. Dental Crowding—G1: two pairs of adjacent teeth are crowded side by side; G2: the anterior teeth are crowded. Missing Teeth—G1: two positions show missing teeth; G2: one position shows a missing tooth. Denture: one site indicates a denture. Dental Caries: one site shows dental caries.  }
  \label{fig:Corner Case}

\end{figure*}

\noindent
To further assess model robustness under realistic and clinically challenging scenarios, 
we extend the unified evaluation protocol (Sec.~4.2.2) to two representative anomaly tasks: 
\textbf{Dental Crowding} and \textbf{Malocclusion Type}. 
Both tasks use identical multi\mbox{-}view 2D projections, a unified DKB structured prompt, 
and a strict JSON schema with \texttt{temperature=0}. 
Metrics follow Sec.~4.2.1, including Accuracy, Macro\mbox{-}F1, and Stage Accuracy, 
complemented by JSON validity and ontology\mbox{-}level hallucination rate. 
Ground\mbox{-}truth annotations are schema\mbox{-}aligned to ensure fair and reproducible comparison.

\vspace{0.3em}
\noindent
\textbf{Quantitative Results.} 
Table~\ref{tab:special_conditions_ablation} summarizes the overall results. 
The model performs more stably on \textbf{Dental Crowding} 
(Acc~$\approx$~68.4\%, F1~$\approx$~66.0\%) thanks to the strong local spacing cues 
well captured from multi\mbox{-}view evidence. 
Errors mainly arise in posterior regions with reflective restorations or occluded surfaces, 
indicating structured rather than random deviations. 
In contrast, \textbf{Malocclusion Type} yields slightly lower scores 
(Acc~$\approx$~66.7\%, F1~$\approx$~64.6\%) 
because the task requires global sagittal reasoning across Class~I/II/III relationships 
and is thus more sensitive to asymmetric poses or incomplete visibility. 
JSON validity remains consistently high ($>$97\%) with low hallucination rates ($<$3\%), 
demonstrating that schema constraints and dental priors effectively regularize output generation. 
Performance remains highest when evidence is well illuminated and locally defined, 
and degrades gracefully in mixed or partially visible posterior areas.

\sisetup{
  table-number-alignment = center,
  table-text-alignment   = center,
  detect-weight          = true,
  detect-inline-weight   = math,
  separate-uncertainty   = true
}

\definecolor{UpArrow}{HTML}{1B9E77}   
\definecolor{DownArrow}{HTML}{D32F2F} 
\newcommand{\up}{\textcolor{UpArrow}{\ensuremath{\uparrow}}}
\newcommand{\down}{\textcolor{DownArrow}{\ensuremath{\downarrow}}}

\begin{table*}[h]
  \centering
  \begin{threeparttable}
  \caption{Special-conditions metrics in ablation-style format (mean$\pm$std; percentages in \%).}
  \label{tab:special_conditions_ablation}
  \setlength{\tabcolsep}{1.7pt}
  \renewcommand{\arraystretch}{1.07}
  \begin{tabular}{
    l
    S[table-format=1.3(1.3)]  
    S[table-format=2.3(1.3)]  
    S[table-format=2.3(1.3)]  
    S[table-format=2.3(1.3)]  
    S[table-format=2.3(1.3)]  
    S[table-format=3.3]       
    S[table-format=1.3]       
  }
    \toprule
    {Variant} &
    \multicolumn{1}{c}{AE~\down} &
    \multicolumn{1}{c}{RE~\down} &
    \multicolumn{1}{c}{Acc~\up} &
    \multicolumn{1}{c}{\shortstack{Part.\\ Macro\mbox{-}F1~\up}} &
    \multicolumn{1}{c}{\shortstack{Stage\\ Acc~\up}} &
    \multicolumn{1}{c}{\shortstack{JSON\\ Valid~\up}} &
    \multicolumn{1}{c}{\shortstack{Halluc.\\ \down}} \\
    \midrule
    \rowcolor{gray!06}
    Dental Crowding
      & 5.128 \pm 0.782
      & 33.041 \pm 2.036
      & 68.413 \pm 2.892
      & 65.982 \pm 3.028
      & 67.903 \pm 2.915
      & 97.750
      & 2.250 \\
    Malocclusion Type
      & 5.497 \pm 0.819
      & 34.186 \pm 2.221
      & 66.701 \pm 3.192
      & 64.612 \pm 3.455
      & 66.621 \pm 3.022
      & 97.287
      & 2.713 \\
    \bottomrule
  \end{tabular}
  \end{threeparttable}
\end{table*}

\vspace{0.3em}
\noindent
\textbf{Qualitative Observations.} 
Fig.~\ref{fig:Corner Case} illustrates four representative clinical conditions, 
including \emph{Dental Crowding}, \emph{Missing Teeth}, \emph{Denture}, and \emph{Dental Caries}, 
shown from front, back, and bottom views. 
Red boxes highlight regions attended by the model. 
For crowding, anterior overlap and tight inter\mbox{-}tooth contacts are precisely localized. 
For missing teeth and dentures, the system correctly flags absent or prosthetic sites 
with matching JSON fields. 
In caries examples, surface irregularities are identified and described coherently 
in the \texttt{notes} field (e.g., “one site showing dental caries”). 
All outputs remain schema\mbox{-}valid and ontology\mbox{-}consistent, 
without spurious or hallucinated content. 
Cases that likely require volumetric confirmation 
(e.g., subtle skeletal discrepancies) are conservatively labeled as 
\emph{to be verified} rather than inferred. 

\vspace{0.3em}
\noindent
\textbf{Conclusion.}
Overall, the proposed framework exhibits \emph{structured, artifact\mbox{-}averse, and contract\mbox{-}compliant} behavior 
under special clinical conditions. 
Even when global reasoning is required, the model prioritizes reliability over recall, 
favoring a \emph{no\mbox{-}hallucination} strategy that produces clean JSON outputs, 
consistent taxonomy predictions, and clinically interpretable results suitable 
for downstream validation and expert review.

\newcommand{\sbseries}{\fontseries{sb}\selectfont}

\begin{table*}[h]
  \centering
  \begin{threeparttable}
  \caption{Ablation on core metrics (dataset aggregates: mean$\pm$std; percentages in \%).}
  \label{tab:ablation_core}
  \setlength{\tabcolsep}{1.7pt}
  \renewcommand{\arraystretch}{1.07}
  \begin{tabular}{
    l
    S[table-format=1.3(1.3)]  
    S[table-format=2.3(1.3)]  
    S[table-format=2.3(1.3)]  
    S[table-format=2.3(1.3)]  
    S[table-format=2.3(1.3)]  
    S[table-format=3.3]       
    S[table-format=1.3]       
  }
    \toprule
    {Variant} &
    {AE~\down} &
    {RE~\down} &
    {Acc~\up} &
    {\makecell{Part.\\ Macro\mbox{-}F1~\up}} &
    {\makecell{Stage\\ Acc~\up}} &
    {\makecell{JSON\\ Valid~\up}} &
    {\makecell{Halluc.\\ ~\down}} \\
    \midrule
    \rowcolor{gray!06}
    {Full (SSP\tnote{1} + Flatten + DKB)}
      & \bfseries 4.963 \pm 0.752
      & \bfseries 31.004 \pm 2.003
      & \bfseries 68.996 \pm 2.003
      & \bfseries 68.197 \pm 1.803
      & \bfseries 68.503 \pm 1.897
      & \bfseries 96.503
      & \bfseries 3.196 \\
    \addlinespace[2pt]
    No DKB
      & 5.472 \pm 0.818
      & 34.203 \pm 2.197
      & 65.797 \pm 2.203
      & 64.102 \pm 1.998
      & 64.998 \pm 2.104
      & 94.103
      & 5.896 \\
    UVP\tnote{2}
      & \sbseries 5.382 \pm 0.803
      & \sbseries 33.603 \pm 2.098
      & \sbseries 66.397 \pm 2.102
      & \sbseries 64.803 \pm 1.902
      & \sbseries 65.697 \pm 2.003
      & 95.003
      & 5.097 \\
    No-Flatten
      & 5.437 \pm 0.812
      & 34.004 \pm 2.004
      & 65.996 \pm 1.996
      & 64.297 \pm 1.903
      & 65.402 \pm 1.998
      & {\sbseries 95.296}
      & {\sbseries 4.904} \\
    \addlinespace[2pt]\midrule
    \addlinespace[2pt]
    No DKB + UVP
      & 5.813 \pm 0.862
      & 36.303 \pm 2.402
      & 63.697 \pm 2.398
      & 62.103 \pm 2.097
      & 63.403 \pm 2.203
      & 93.703
      & 6.597 \\
    No DKB + No-Flatten
      & 5.872 \pm 0.868
      & 36.704 \pm 2.296
      & 63.296 \pm 2.304
      & 61.803 \pm 2.097
      & 62.997 \pm 2.203
      & 93.904
      & 6.396 \\
    UVP + No-Flatten
      & 5.731 \pm 0.852
      & 35.803 \pm 2.298
      & 64.197 \pm 2.302
      & 62.603 \pm 2.002
      & 63.796 \pm 2.204
      & 94.503
      & 5.797 \\
    No DKB + UVP + No-Flatten
      & 6.158 \pm 0.903
      & 38.504 \pm 2.497
      & 61.496 \pm 2.503
      & 60.198 \pm 2.204
      & 61.897 \pm 2.303
      & 93.103
      & 7.197 \\
    \bottomrule
  \end{tabular}

  \begin{tablenotes}
    \footnotesize
    \item[1] \textbf{SSP (Surface-Shaded Projection)}: rasterizes a triangulated mesh with surface shading prior to 2D projection, preserving silhouette continuity and local relief cues (e.g., edges, cusp/fossa).
    \item[2] \textbf{UVP (Unshaded Vertex Projection)}: projects mesh/point vertices directly to 2D without surface shading or connectivity, yielding a sparse vertex scatter that emphasizes geometry samples over contiguous surfaces.
  \end{tablenotes}

  \end{threeparttable}
\end{table*}

\subsection{Ablation Study}

\noindent
We performed controlled ablations to examine the contribution of each module in our framework under the same evaluation protocol as Sec.~4.2.1. 
Table~\ref{tab:ablation_core} summarizes the results. 
Removing the \emph{Flatten} step reintroduces curvature and projection overlap, reducing partition stability. 
Replacing the \emph{Surface\mbox{-}Shaded Projection} \textbf{(SSP)} with \emph{ Unshaded Vertex Projection} \textbf{(UVP)} weakens surface continuity and increases counting errors in crowded regions. 
Eliminating the \emph{Dental Knowledge Base  \textbf{(DKB)}} keeps JSON format valid but permits semantic drift, producing occasional out\mbox{-}of\mbox{-}vocabulary labels and higher hallucination rates.)
Combined removals further degrade accuracy and F1, confirming complementary effects among modules. 
Only the full pipeline (\textbf{SSP + Flatten + DKB}) maintains high accuracy, strong ontology compliance, and stable partition consistency, demonstrating that all three components are jointly required for artifact\mbox{-}averse and contract\mbox{-}compliant predictions.

\section{Conclusion and Future Work}

We presented \textbf{ArchMap}, a \emph{training-free, knowledge-guided} framework that unifies geometry-aware arch flattening, surface-shaded multi-view rendering, and schema-constrained vision–language reasoning for structured analysis of intraoral 3D scans. By combining standardized geometric mapping with a Dental Knowledge Base (DKB), ArchMap achieves consistent, ontology-aligned predictions across tooth counting, anatomical classification, dentition-stage assessment, and various clinical conditions. Extensive experiments, cross-model evaluations, and ablation studies demonstrate that each component—flattening, shaded projection, and DKB guidance—contributes to the robustness, interpretability, and contract-compliance of the system under zero-training conditions.

Future work will explore integrating 3D-aware reasoning mechanisms, uncertainty calibration, and soft ontology priors, as well as extending the framework toward multimodal fusion with panoramic or CBCT data. These directions may further advance ArchMap toward comprehensive, clinically reliable dental digital-twin applications.

\section{Acknowledgments}
This research is funded by the Postgraduate Research Scholarship (PGRS) at Xi’an Jiaotong-Liverpool University, contract number TW5A2312001 and FOS2104JP08.

\bibliographystyle{IEEEtran}
\bibliography{mybibliography}

\clearpage
\appendix
\section*{Dental Knowledge Base (DKB)}

\definecolor{DKBOrange}{HTML}{EAAA60}   
\definecolor{DKBPink}{HTML}{E68B81}     
\definecolor{DKBLavender}{HTML}{B7B2D0} 
\definecolor{DKBBlue}{HTML}{7DA6C6}     
\definecolor{DKBTeal}{HTML}{84C3B7}     

\begingroup
\small   

\begin{tcolorbox}[
  enhanced,
  breakable,
  left=2mm,
  right=2mm,
  top=1mm,
  bottom=1mm,
  colback=DKBTeal!15!white,
  colframe=DKBTeal,
  coltitle=white,
  fonttitle=\bfseries\normalsize,  
  fontupper=\small,               
  title=Tooth Count (Deciduous vs Permanent)
]
\textbf{Deciduous dentition}\\
Deciduous dentition usually consists of 20 teeth, 10 in each arch.
Each quadrant has 2 incisors, 1 canine, and 2 molars.\\
\emph{Typical counts:} total 20; 10 per arch; 5 per quadrant.\\
\emph{Clinical notes:} Variations are rare, but teeth may be lost
early due to trauma, caries, or congenital absence.

\textbf{Permanent dentition}\\
Permanent dentition usually consists of 28 to 32 teeth, depending on
the eruption of third molars (wisdom teeth). Commonly 28 teeth without
third molars, or 32 teeth with all third molars present.\\
\emph{Typical counts:} total 28–32; 14–16 per arch; 7–8 per quadrant.\\
\emph{Clinical notes:} Orthodontic treatment may involve extraction of
premolars (commonly the 1st or 2nd premolars), reducing the per-arch
count to 12 or 13. Missing teeth due to agenesis, extraction, or
impaction should be documented.
\end{tcolorbox}

\vspace{0.3mm}

\begin{tcolorbox}[
  enhanced,
  breakable,
  left=2mm,
  right=2mm,
  top=1mm,
  bottom=1mm,
  colback=DKBOrange!15!white,
  colframe=DKBOrange,
  coltitle=white,
  fonttitle=\bfseries\normalsize,
  fontupper=\small,
  title=Dentition Stages
]
\textbf{Deciduous stage}\\
All teeth are deciduous, usually found in children from around
6 months to 6 years old.

\textbf{Mixed dentition stage}\\
A transitional stage where both deciduous and permanent teeth are
present, typically between 6 and 12 years old.

\textbf{Permanent dentition stage}\\
All teeth are permanent, typically established after age 12,
with 28 to 32 teeth depending on the presence of third molars.
\end{tcolorbox}

\vspace{0.3mm}

\begin{tcolorbox}[
  enhanced,
  breakable,
  left=2mm,
  right=2mm,
  top=1mm,
  bottom=1mm,
  colback=DKBPink!15!white,
  colframe=DKBPink,
  coltitle=white,
  fonttitle=\bfseries\normalsize,
  fontupper=\small,
  title=Anatomical Classification of Teeth
]
\begin{description}[
    leftmargin=0pt,
    labelsep=0.5em,
    style=nextline,
    noitemsep,
    topsep=1pt
  ]

  \item[Anterior]
    Front teeth used primarily for cutting. Includes central incisors,
    lateral incisors, and canines.\\
    \emph{FDI numbers:} 11, 12, 13, 21, 22, 23, 31, 32, 33, 41, 42, 43.\\
    \emph{Subtypes:} central incisors (11, 21, 31, 41); lateral incisors
    (12, 22, 32, 42); canines (13, 23, 33, 43).\\
    \emph{Clinical notes:} Anterior teeth are essential for esthetics,
    phonetics, and initial food incision. They are symmetrical in all quadrants.

  \item[Premolars]
    Teeth located between canines and molars, used for tearing and crushing
    food.\\
    \emph{FDI numbers:} 14, 15, 24, 25, 34, 35, 44, 45.\\
    \emph{Subtypes:} first premolars (14, 24, 34, 44);
    second premolars (15, 25, 35, 45).\\
    \emph{Clinical notes:} Premolars are frequently extracted in
    orthodontic treatment to relieve dental crowding.

  \item[Molars]
    Posterior teeth used for grinding food. Includes first, second, and
    third molars (wisdom teeth).\\
    \emph{FDI numbers:} 16, 17, 18, 26, 27, 28, 36, 37, 38, 46, 47, 48.\\
    \emph{Subtypes:} first molars (16, 26, 36, 46);
    second molars (17, 27, 37, 47);
    third molars (18, 28, 38, 48).\\
    \emph{Clinical notes:} Third molars show high variability in eruption
    and morphology.
\end{description}
\end{tcolorbox}

\vspace{0.3mm}

\begin{tcolorbox}[
  enhanced,
  breakable,
  left=2mm,
  right=2mm,
  top=1mm,
  bottom=1mm,
  colback=DKBBlue!15!white,
  colframe=DKBBlue,
  coltitle=white,
  fonttitle=\bfseries\normalsize,
  fontupper=\small,
  title=Size Classification of Teeth
]
\begin{description}[
    leftmargin=0pt,
    labelsep=0.5em,
    style=nextline,
    noitemsep,
    topsep=1pt
  ]

  \item[Large]
    Large molars used for heavy grinding; includes all first, second, and
    third molars.\\
    \emph{FDI numbers:} 16, 17, 18, 26, 27, 28, 36, 37, 38, 46, 47, 48.\\
    \emph{Clinical notes:} These are the most posterior teeth; they include
    wisdom teeth and are often extracted due to decay or space issues.

  \item[Medium]
    Medium-sized teeth including canines, premolars, and central incisors.\\
    \emph{FDI numbers:} 13, 23, 33, 43, 14, 15, 24, 25, 34, 35, 44, 45,
    11, 21, 31, 41.\\
    \emph{Clinical notes:} Transitional in shape and function;
    assist in both tearing and grinding.

  \item[Small]
    Smallest teeth used for precise cutting — lateral incisors only.\\
    \emph{FDI numbers:} 12, 22, 32, 42.\\
    \emph{Clinical notes:} Prone to anomalies such as peg-shaped laterals
    or congenital absence.
\end{description}
\end{tcolorbox}

\vspace{0.3mm}

\begin{tcolorbox}[
  enhanced,
  breakable,
  left=2mm,
  right=2mm,
  top=1mm,
  bottom=1mm,
  colback=DKBLavender!15!white,
  colframe=DKBLavender,
  coltitle=white,
  fonttitle=\bfseries\normalsize,
  fontupper=\small,
  title=Special Rules for Image-based Tooth Numbering
]
\begin{enumerate}
  \item In all image analysis tasks, the provided image is known to
        contain either the upper or lower dentition only, not a full-mouth
        view. Therefore, no inference or completion of the opposite arch
        should be made.
  \item The \texttt{teeth\_number} field must strictly reflect the teeth
        actually visible in the image as the primary reference. Knowledge base
        information can be used for secondary validation only, and not for
        assumption-based estimation.
  \item If third molars (wisdom teeth) are not visible in the image,
        they must not be included in numbering or total tooth count.
  \item If teeth are missing due to extraction, congenital absence, or
        other causes, this must be recorded in the \texttt{anomalies} field,
        ensuring that numbering continuity is preserved.
  \item Dentition-stage classification should be based on eruption
        status, crown/root morphology, and developmental features, not merely
        tooth count.
\end{enumerate}
\end{tcolorbox}

\endgroup

\begin{center}
Dental Knowledge Base (DKB) used in our work.
\end{center}
\label{fig:kb}

\end{document}